\documentclass[sigconf,nonacm,balance=false]{acmart}
\setcopyright{none}
\makeatletter
\renewcommand{\@ACM@checkaffil}{}
\makeatother
\usepackage{graphicx}
\usepackage{amsmath}
\usepackage{booktabs}
\usepackage{tabularx}
\usepackage{array}
\usepackage{xcolor}
\usepackage[above,below]{placeins}
\definecolor{bestred}{RGB}{210,20,32}
\definecolor{secondblue}{RGB}{24,92,178}
\DeclareRobustCommand{\best}[1]{\textcolor{bestred}{\textbf{#1}}}
\DeclareRobustCommand{\second}[1]{\textcolor{secondblue}{\textbf{#1}}}

\newsavebox{\comparisonbox}
\newenvironment{fitcolumn}{\begin{lrbox}{\comparisonbox}}{%
\end{lrbox}%
\ifdim\wd\comparisonbox>\linewidth
\resizebox{\linewidth}{!}{\usebox{\comparisonbox}}%
\else\usebox{\comparisonbox}\fi}

\title{Foresee to Trade: Forecast-Augmented Policy Optimization for Portfolio Management}
\author{Zeyu LIU}
\authornote{Equal contribution.}
\author{Xuanzhi FENG}
\authornotemark[1]
\author{Haoran XU}
\authornotemark[1]
\author{LAI Sing Kwong}
\authornotemark[1]
\author{Yang LI}
\author{Hualei ZHANG}
\author{Hao CHEN}
\author{Jie ZHANG}
\author{Xiaoyi PANG}
\authornote{Corresponding authors.}
\author{Song GUO}
\authornotemark[2]
\affiliation{\institution{The Hong Kong University of Science and Technology}}
\email{xypang@whu.edu.cn}
\email{songguo@ust.hk}
\renewcommand{\shortauthors}{Zeyu LIU et al.}
\begin{document}

\begin{abstract}
Online investment services must translate asset opportunities into investment outcomes for users' accounts, yet the same market opportunity can warrant different account actions. We study how supervision from future asset opportunities can strengthen account-level trading decisions and propose Forecast-Augmented Relative Policy Optimization (FRPO). FRPO separately supervises asset opportunities and account decisions through a jointly learned representation: a Financial State Head (FSH) learns future relative-return targets to provide asset-level opportunity supervision, while Same-Path Group-Relative Optimization compares trading trajectories under the same market path and initial account to provide account-level relative-outcome supervision. An Adaptive Gate coordinates auxiliary supervision using predictive variance and gradient alignment. FSH is used only during training; inference uses the encoder and policy for trading. Across DJIA30, TOPIX Core30, and three investment-style portfolios, ten-seed evaluations show that FRPO achieves the highest mean cumulative and annualized returns among the compared methods in all five settings. These results support strengthening investment agents' trading decisions through joint supervision from asset opportunities and trading outcomes.
\end{abstract}

\maketitle

\section{Introduction}

Online investment services help users discover opportunities, allocate
capital, and rebalance portfolios~\citep{zhang2024finagent,yu2024fincon,zhang2024earnmore}.
Investment agents must translate market information into trading actions that improve
account outcomes under existing holdings, available cash, and transaction
costs. Connecting asset opportunities with account decisions is therefore
central to learning for agent-mediated investment services.

Financial reinforcement learning trains policies on portfolio
outcomes~\citep{liu2021finrl,liu2022finrlmeta,wang2021deeptrader}.
Predictive learning in portfolio selection and market making additionally
exploits market regularities~\citep{choi2024deepclair,niu2024imm}.
This direction raises the question of how opportunity learning and
account-decision learning should interact during policy training.

\begin{figure*}[t]
\centering
\includegraphics[width=\textwidth]{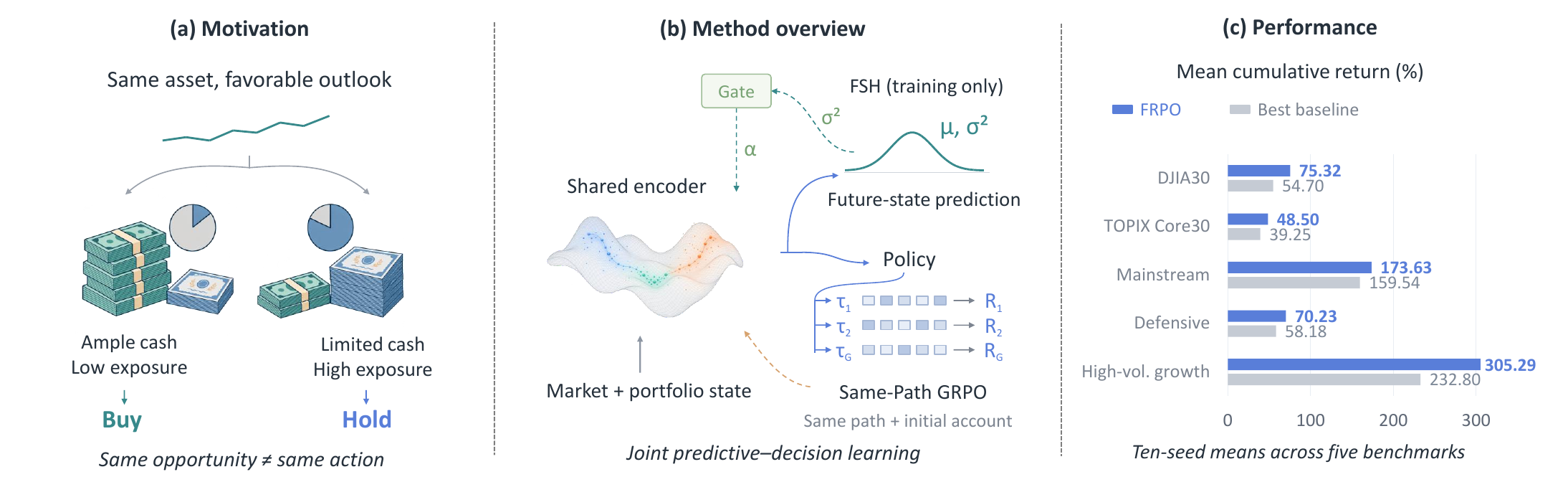}
\caption{From asset opportunities to portfolio actions.
(a) Illustrative accounts link the same favorable asset
outlook to different actions under different cash balances and exposures.
(b) Same-path relative outcomes and future-state targets jointly supervise
the shared encoder; predictive variance and gradient alignment weight auxiliary learning.
The account and representation artwork is schematic.
(c) Ten-seed mean cumulative returns exceed the strongest non-FRPO method
in each setting (Tables~\ref{tab:large_markets}
and~\ref{tab:heterogeneous_portfolios}). Solid arrows show forward flow;
dashed arrows show training supervision.}
\Description{Three panels connect motivation, method, and performance. The illustrative accounts share a favorable Asset A outlook; ample cash and low exposure accompany a Buy example, while limited cash and high exposure accompany Hold. FRPO uses a shared encoder with separate policy and training-only FSH branches. Same-path relative returns and future-state targets supervise the encoder; FSH predictive variance and gradient alignment control the auxiliary weight. A common-scale horizontal bar chart compares ten-seed mean cumulative returns against the strongest non-FRPO method on DJIA30, TOPIX Core30, Mainstream, Defensive, and High-Volatility Growth.}
\label{fig:motivation}
\end{figure*}

Under the price-taker assumption, future prices evolve independently of an
agent's trades, determining relative asset opportunities. Account outcomes
depend on actions that change cash, holdings, and subsequent trading
possibilities. A favorable outlook may support buying for an account with
ample cash and low exposure, but holding for one with substantial exposure.
Thus, the same opportunity can correspond to different actions and investment
outcomes. \emph{How can supervision from future asset opportunities strengthen
account-level trading decisions?}

We propose Forecast-Augmented Relative Policy Optimization (FRPO).\footnote{We will release code, experimental configurations, reproduction scripts, and analysis and visualization scripts (Appendix~\ref{app:open-source-resources}).}
FRPO separately supervises asset opportunities and account decisions through
a jointly learned representation. A shared encoder processes market and
account observations, supporting a portfolio policy and a Financial State
Head (FSH). Opportunity supervision and relative trading-outcome supervision
jointly shape the representation used for trading.

FSH learns future relative-return targets from historical training windows,
with cross-sectional centering, historical-volatility normalization, and
clipping. For account decisions, Same-Path Group-Relative Optimization samples
trajectories from an identical initial account over the same recorded market
path. Different actions produce different holdings, cash balances, and
returns. Comparing these returns provides relative-outcome supervision under
common market conditions, without a separately learned value network.

An Adaptive Gate weights opportunity supervision using FSH predictive
variance and gradient alignment with the policy objective. At inference,
the encoder and policy use available observations; FSH predictions are not
passed to the policy head. Figure~\ref{fig:motivation} summarizes this joint
supervision.

Across DJIA30, TOPIX Core30, and three investment-style portfolios, ten-seed
evaluations show that FRPO achieves the highest mean cumulative and
annualized returns among the compared methods in all five settings.

Our contributions are threefold:
\begingroup
\renewcommand{\labelenumi}{\arabic{enumi})}
\setlength{\leftmargini}{1.5em}
\begin{enumerate}
\setlength{\itemsep}{2pt}
\setlength{\parskip}{0pt}
\setlength{\parsep}{0pt}
\item We introduce FRPO, connecting asset-opportunity and account-decision
learning through separate supervision and an adaptively weighted shared
representation.
\item We develop critic-free group-relative learning for continuous portfolio
trading under shared market paths and initial accounts, and analyze when
these comparisons reduce common market factors' influence on relative
outcomes.
\item We demonstrate return advantages across two equity universes and three
investment-style portfolios with ten seeds. Component ablations, matched
decision-supervision controls, cost sensitivity, and shifted temporal
evaluations examine the mechanisms and applicability of the training design.
\end{enumerate}
\endgroup

\section{Related Work}

\subsection{Web Financial Services and Portfolio Decisions}
Financial information systems support opportunity discovery. FinGPT adapts
language models to financial data~\citep{yang2023fingpt}, and FAST models
time-sensitive news and discussions~\citep{sawhney2021fast}. FinAgent combines
market information, tools, and reflection for trading~\citep{zhang2024finagent};
FinCon~\citep{yu2024fincon} and FinThink~\citep{liu2025finthink} study
multi-agent financial reasoning.

Web research also directly addresses portfolio actions. EarnMore (WWW 2024)
learns maskable representations for customizable
stock pools~\citep{zhang2024earnmore}; HedgeAgents (WWW 2025 Companion)
coordinates a fund manager and asset-class experts for
hedging~\citep{li2025hedgeagents}. Agent Market Arena evaluates trading agents
on live stock and cryptocurrency data using simulated
portfolios~\citep{qian2025ama}.

These studies address information processing, investable universes, agent
coordination, and evaluation. FRPO strengthens trading decisions through
supervision connecting asset opportunities and account trading outcomes.

\subsection{Financial RL and Predictive Supervision}
Rule-based methods use technical indicators and statistical regularities
to make trading decisions~\citep{murphy1999technical,kakushadze2016formulaic}.
Financial RL platforms~\citep{liu2021finrl,liu2022finrlmeta} support
PPO~\citep{schulman2017ppo}, A2C~\citep{mnih2016asynchronous},
TD3~\citep{fujimoto2018td3}, and SAC~\citep{haarnoja2018sac}.
Portfolio research explores policy designs including DeepTrader's market
embedding~\citep{wang2021deeptrader}, MASA's multi-agent risk
control~\citep{li2024masa}, DeepAries' rebalancing
intervals~\citep{kim2025deeparies}, and ArchetypeTrader's strategy
archetypes~\citep{zong2026archetypetrader}. For policy supervision, Financial
DAPO combines group-relative updates with news-derived
signals~\citep{zha2025dapo}.

Predictive learning also strengthens financial decisions. DeepClair uses
forecast outputs for portfolio allocation~\citep{choi2024deepclair}, while IMM
uses predictive signals as policy inputs and for expert
guidance~\citep{niu2024imm}. Through joint supervision from future asset
opportunities and same-path relative trading outcomes, FRPO integrates
opportunity learning into the shared representation to strengthen trading decisions.

\section{Methodology}

\subsection{FRPO Framework}
\label{sec:framework}
\begin{figure*}[t]
\centering
\includegraphics[page=1,width=\textwidth]{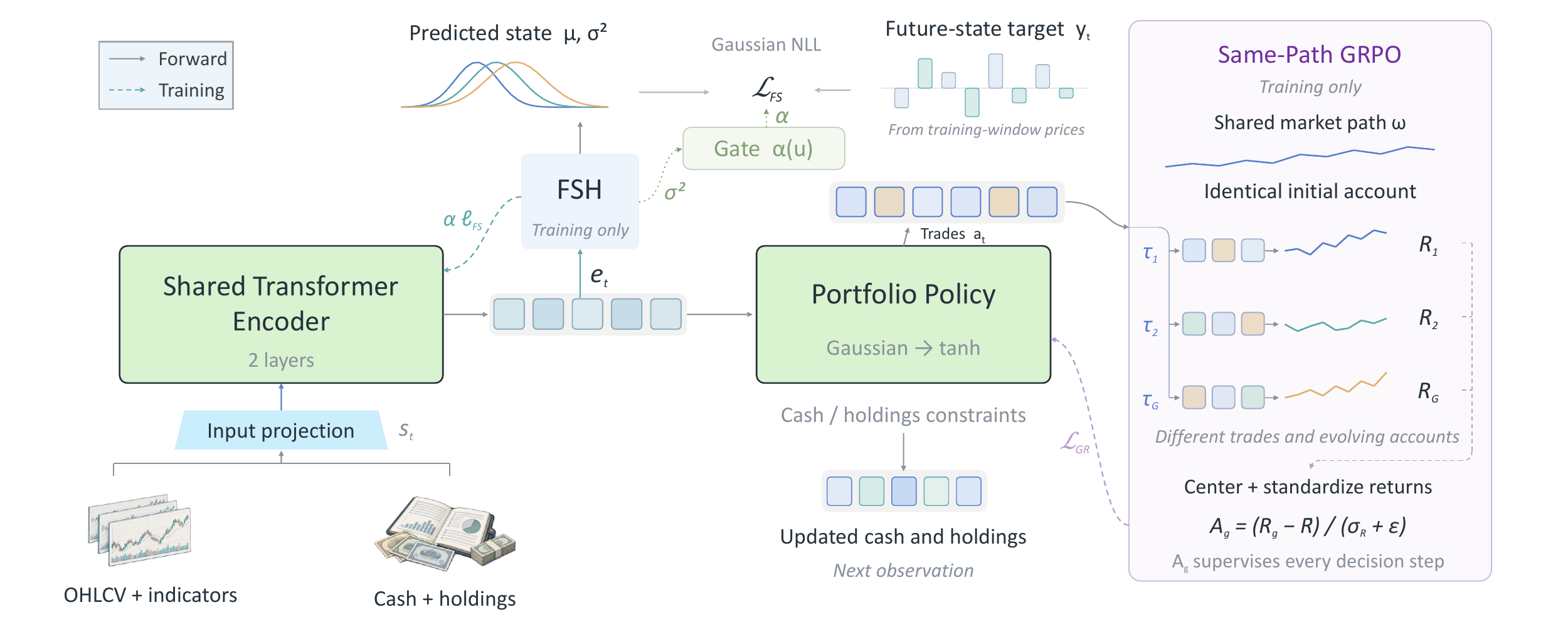}
\caption{FRPO architecture with shared predictive and decision supervision.
Market observations and cash/holdings enter one encoder. Its representation
supports a continuous portfolio policy and the upper Financial State Head
(FSH). Centered, historical-volatility-normalized, clipped future-return
targets supervise FSH through heteroscedastic Gaussian NLL; predictive
variance and gradient alignment determine a detached auxiliary weight. The right inset compares
full-horizon net returns on a shared market path and initial account, while
later accounts diverge. Each group-relative advantage supervises every step
of its trajectory. Solid arrows show forward flow, dashed arrows training,
and dotted arrows loss weighting. Input artwork and trajectory traces are
schematic. Inference uses the encoder and policy alone.}
\Description{An open horizontal diagram shows illustrated historical market charts and a current account ledger merging before input projection and a shared Transformer encoder. Shared features continue right into a continuous portfolio policy and branch upward into a training-only Financial State Head. Predicted distributions and future-state targets meet at a Gaussian NLL objective. Predictive variance and gradient alignment determine the detached auxiliary weight. Short dashed curves show auxiliary supervision of the encoder and group-relative supervision of the policy. A narrow right inset contains one shared external price path and identical initial account, three different trading and account trajectories, and all three full-horizon returns collected for group-relative advantages. Each advantage applies at every decision step. Input illustrations and trajectory traces are conceptual, not measured experimental results. FSH predictions do not enter the policy.}
\label{fig:framework}
\end{figure*}

FRPO learns a portfolio policy from asset-level supervision of future
relative opportunities and portfolio-level supervision of trading outcomes
(Figure~\ref{fig:framework}). Observation $\mathbf s_t$ contains historical
market features and the current account state, including cash and holdings.
Under the price-taker assumption, the market is exogenous; the account evolves
with trading actions and determines feasible reallocations.
The shared encoder $\mathbf e_t=f_\phi(\mathbf s_t)\in\mathbb R^{64}$ supplies the policy
$\mathbf a_t\sim\pi_\theta(\cdot\mid\mathbf e_t)$ and auxiliary predictor
$p_\psi(\mathbf y_t\mid\mathbf e_t)$.
The horizontal path executes account-feasible trading actions. The upper auxiliary
branch supervises FSH with targets derived from future training-window prices.
The right inset compares complete trading trajectories under a shared market
path and initial account. The Adaptive Gate weights the auxiliary contribution
to the shared encoder.
FSH predictions are not inputs to the policy head. At inference, the encoder
and policy use available observations alone.

\subsection{Same-Path Group-Relative Optimization}
\label{sec:samepath}

\textbf{Why same-path comparisons?}
Portfolio returns reflect both allocation decisions and external market
movements. Their short-horizon statistical behavior motivates evaluating
candidate actions under common market conditions.

Let $x_t=\log p_t$ denote the log price of an asset. Under an
Euler--Maruyama discretization of an It{\^o} diffusion~\citep{ito1944stochastic,
maruyama1955continuous}, its short-horizon increment can be approximated as
\begin{equation}
\Delta x_t=\mu_t\Delta t+\sigma_t\Delta W_t,
\qquad
\Delta W_t=\sqrt{\Delta t}\,\varepsilon_t.
\label{eq:em}
\end{equation}
For locally bounded drift and nonzero finite diffusion, the predictable
increment in Equation~\ref{eq:em} is $O(\Delta t)$ and the standard deviation
of stochastic variation is $O(\sqrt{\Delta t})$. Their ratio is
$O(\sqrt{\Delta t})$: shortening the interval weakens the drift signal relative
to market randomness, making short-horizon realized returns noisy feedback
for trading decisions.

Same-path comparison is related to the common-random-numbers principle in
simulation: alternatives share an exogenous realization as their evaluation
condition~\citep{glasserman1992crn}. Different actions still create different
holdings and market exposures. The effect on return differences depends on
the covariance induced by the shared path; it does not guarantee lower total
policy-gradient variance.

We adapt group-relative optimization~\citep{shao2024deepseekmath} by sampling
$G$ trajectories on a fixed historical segment $\omega$. Each trajectory starts
with identical cash and holdings and follows the same price sequence under
the old policy:
\begin{equation}
\tau_\omega^{(g)}
\sim
\pi_{\theta_{\mathrm{old}}}(\cdot\mid\omega),
\qquad
g=1,\ldots,G.
\label{eq:rollout}
\end{equation}
The trajectories in Equation~\ref{eq:rollout} share the external price path
but develop different actions, cash balances, and holdings. Trajectory $g$ receives
the cumulative net trading reward
$R_\omega^{(g)}=\sum_t\gamma^t r_{\omega,t}^{(g)}$, where $\gamma=1$ and
$r_{\omega,t}^{(g)}$ is $10^{-4}$ times the account NAV increment after transaction fees.

We center and standardize the trajectory returns within the group to obtain a
single relative advantage shared by the entire trajectory:
\begin{equation}
\begin{aligned}
A_\omega^{(g)}
&=
\frac{R_\omega^{(g)}-\overline R_\omega}
{\sigma_{R,\omega}+\epsilon},
\qquad
\overline R_\omega
=
\frac1G\sum_{j=1}^G R_\omega^{(j)},\\
\sigma_{R,\omega}^2
&=
\frac1G\sum_{j=1}^G
\left(R_\omega^{(j)}-\overline R_\omega\right)^2,
\end{aligned}
\label{eq:groupadv}
\end{equation}
where $\epsilon>0$ is a numerical stability constant. Centering focuses the
policy update on relative performance among trajectories under the same market
condition, while standardization mitigates return-scale differences across
market periods and trajectory groups.

An identical additive return component cancels under group centering.
With non-anticipating dynamic holdings, residual factor-innovation variance
equals risk-weighted exposure dispersion. Same-path comparison reduces
this component when dispersion is small relative to common exposure risk.
Appendix~\ref{app:same-path-factor} derives the expectation-level comparison
before standardization.
Effective exposure uses executed holdings and reward scaling,
incorporating cash limits, transaction costs, and account
feedback. Trading may respond to past prices and portfolio changes but
must precede the next market innovation. The comparison concerns the
factor-innovation part of returns; other components can remain correlated
with it.
The trajectory-level advantage in Equation~\ref{eq:groupadv} is broadcast to
all rebalancing steps, adapting sequence-level GRPO supervision to trading.

The continuous rebalancing policy $\pi_\theta$ is Gaussian. Samples pass through
$\tanh$ and account constraints to form executed trading actions $\mathbf a_t^{\mathrm{exec}}$:
sells are capped by holdings and buys by available cash, precluding short selling and leverage.
Each action therefore changes both immediate exposure and the reallocations
available at subsequent steps. Transaction costs enter the realized trading
outcome. Cash remains a residual balance with no independent action dimension.

Define the new-to-old policy density ratio and its clipped form as
\begin{equation}
\begin{aligned}
\rho_{\omega,t}^{(g)}
&=
\frac{
\pi_\theta(\mathbf a_{\omega,t}^{(g)}\mid\mathbf s_{\omega,t}^{(g)})
}{
\pi_{\theta_{\mathrm{old}}}
(\mathbf a_{\omega,t}^{(g)}\mid\mathbf s_{\omega,t}^{(g)})
},
\\
\widehat\rho_{\omega,t}^{(g)}
&=
\operatorname{clip}\!\left(
\rho_{\omega,t}^{(g)},1-\varepsilon_c,1+\varepsilon_c
\right),
\end{aligned}
\label{eq:ratio}
\end{equation}
In Equation~\ref{eq:ratio}, $\varepsilon_c$ is the policy clipping coefficient. We denote the local
group-relative policy loss for trajectory $g$ at time $t$ by
$\ell_{\mathrm{GR},\omega,t}^{(g)}$:
\begin{equation}
\ell_{\mathrm{GR},\omega,t}^{(g)}
=
-\min\!\left(
\rho_{\omega,t}^{(g)}A_\omega^{(g)},
\widehat\rho_{\omega,t}^{(g)}A_\omega^{(g)}
\right),
\label{eq:grlocal}
\end{equation}
The expectation of Equation~\ref{eq:grlocal} defines the overall loss:
\begin{equation}
\mathcal L_{\mathrm{GR}}
=
\mathbb E_{\omega,g,t}
\left[
\ell_{\mathrm{GR},\omega,t}^{(g)}
\right],
\label{eq:grpo}
\end{equation}
In Equation~\ref{eq:grpo}, the expectation is over market segments,
trajectories, and steps. The update uses relative outcomes without a separate
value function or generalized advantage estimator~\citep{schulman2017ppo,shao2024deepseekmath}.

Relative advantages supervise the policy through trajectory outcomes.
Future-state targets provide complementary asset-level supervision for the
shared encoder.
\subsection{Financial State Modeling}

The Financial State Head provides supervision for future cross-asset
opportunities. It models future relative-return states and their predictive
variance, with $\omega$ and $g$ suppressed below.
For asset $i$, define
the forward log return as
$u_{t,i}=\log(p_{t+h_t,i}/p_{t,i})$, where
$h_t=\min(H,T-t)$ for $t<T$, $H=5$ trading days, $T$ is the final observed
time of the sampled rollout window, and $N$ is the number of assets.
The lookahead is fixed at $H$ whenever available and shortened only near
the rollout boundary, retaining end-of-window training targets without
accessing prices beyond $T$. We remove the contemporaneous market-wide direction by cross-sectional
centering,
$\widetilde u_{t,i}=u_{t,i}-N^{-1}\sum_{j=1}^{N}u_{t,j}$, and then normalize by
historical volatility to construct a risk-adjusted Financial State target:
\begin{equation}
y_{t,i}
=
\operatorname{clip}\!\left(
\frac{\widetilde u_{t,i}}{\nu_{t,i}+\epsilon},
-y_{\max},y_{\max}
\right).
\label{eq:statetarget}
\end{equation}
Here, $\nu_{t,i}$ is the historical volatility of asset $i$ at time $t$, and
$y_{\max}$ is the target clipping threshold.

FSH models the conditional distribution of the target in
Equation~\ref{eq:statetarget}. From $\mathbf e_t$, it outputs a conditional
mean and a heteroscedastic predictive variance:
\begin{equation}
\left(
\mu_{t,i}^{\mathrm{FS}},
\log\sigma_{t,i}^{2,\mathrm{FS}}
\right)
=
g_{\psi,i}(\mathbf e_t).
\label{eq:fshead}
\end{equation}
In Equation~\ref{eq:fshead}, the mean predicts the centered, normalized,
and clipped target, and the variance models its residual variability given
the current observations.

The per-step Financial State loss uses twice the heteroscedastic Gaussian
negative log-likelihood, with its constant term omitted:
\begin{equation}
\ell_{\mathrm{FS},t}
=
\frac1N\sum_{i=1}^{N}
\left[
\frac{
\left(y_{t,i}-\mu_{t,i}^{\mathrm{FS}}\right)^2
}{
\sigma_{t,i}^{2,\mathrm{FS}}
}
+\log\sigma_{t,i}^{2,\mathrm{FS}}
\right].
\label{eq:fs}
\end{equation}
The expectation of Equation~\ref{eq:fs} gives
$\mathcal L_{\mathrm{FS}}=\mathbb E_t[\ell_{\mathrm{FS},t}]$. The gate below
reduces the auxiliary weight for targets with larger predicted variance.

\subsection{Adaptive Gate}

The Adaptive Gate adjusts the contribution of future-state supervision
using FSH predictive variance and gradient alignment. In optimization minibatch $k$, let $j$
index a trajectory--time sample and define
$u_{k,j}=N^{-1}\sum_i\sigma_{k,j,i}^{2,\mathrm{FS}}$.
The uncertainty scale $\tau$ is the median $u_{1,j}$ in the first gated
minibatch, floored at $10^{-8}$ and then held fixed throughout training.
Let $c_k$ be the cosine similarity between the clipped policy-loss gradient
and the unweighted FSH-loss gradient with respect to the shared encoder
sequence features of the entire minibatch. We maintain $m_1=c_1$ and
$m_k=0.9m_{k-1}+0.1c_k$, and set
\begin{equation}
g_k^{\mathrm{align}}=
\operatorname{clip}\!\left(\frac{1+m_k}{2},\,0.1,\,1\right).
\label{eq:aligngate}
\end{equation}
The adaptive weight for sample $j$ is
\begin{equation}
\alpha_{k,j}
=
\operatorname{sg}\!\left[
\alpha_0\exp(-u_{k,j}/\tau)g_k^{\mathrm{align}}
\right],
\label{eq:alphagate}
\end{equation}
where $\alpha_0$ is the base weight and $\operatorname{sg}[\cdot]$ denotes
the stop-gradient operator. This gives $\alpha_{\omega,t}^{(g)}$ for the
corresponding sample in the joint objective. The alignment factor is shared
within each minibatch, and its moving average persists across updates.
Stop-gradient prevents backpropagation through the gate weight.

\subsection{Joint Training Objective}

The complete FRPO objective is
\begin{equation}
\begin{aligned}
\mathcal L_{\mathrm{FRPO}}
&=
\mathcal L_{\mathrm{GR}}
-\lambda_{\mathcal H}
\mathbb E_{\omega,g,t}
\left[
\mathcal H\!\left(
\pi_\theta(\cdot\mid\mathbf s_{\omega,t}^{(g)})
\right)
\right]\\
&\quad
+\mathbb E_{\omega,g,t}
\left[
\alpha_{\omega,t}^{(g)}
\ell_{\mathrm{FS},\omega,t}^{(g)}
\right],
\end{aligned}
\label{eq:joint}
\end{equation}
where $\mathcal H(\pi_\theta)$ denotes policy entropy and
$\lambda_{\mathcal H}\geq0$ is the entropy-regularization coefficient. Because
the objective is minimized, the negative entropy term encourages exploration.

Equation~\ref{eq:joint} trains predictive and decision features in the same
encoder. The group-relative loss rewards trading outcomes, the Financial State
loss supervises future relative opportunities, and the gate in
Equation~\ref{eq:alphagate} controls the auxiliary contribution.

\section{Experiments}

\begin{table*}[!t]
\caption{Performance on two large market universes. Technical and market baselines are deterministic price-only backtests; learned methods report ten-seed averages. CR and AR are cumulative and annualized returns (\%); higher is better for all metrics. \best{Red} and \second{blue} denote the best and second-best values in each dataset and metric.}
\label{tab:large_markets}
\centering
{
\fontsize{8.5}{10.2}\selectfont
\setlength{\tabcolsep}{3pt}
\renewcommand{\arraystretch}{1.14}
\begin{tabular*}{\linewidth}{@{\extracolsep{\fill}}l*{5}{r}@{\hspace{6pt}}*{5}{r}@{}}
\toprule
Method & \multicolumn{5}{c}{DJIA30} & \multicolumn{5}{c}{TOPIX Core30}\\
\cmidrule(lr){2-6} \cmidrule(lr){7-11}
 & CR (\%) & AR (\%) & Sharpe & Sortino & Calmar & CR (\%) & AR (\%) & Sharpe & Sortino & Calmar\\
\midrule
\multicolumn{11}{@{}l}{\textit{Technical and Market Baselines}}\\
MACD & -8.47 & -3.27 & -0.200 & -0.160 & -0.100 & 19.23 & 7.13 & 0.570 & 0.550 & \second{0.660}\\
SMA & 5.63 & 2.08 & 0.110 & 0.090 & 0.060 & -2.36 & -0.93 & -0.070 & -0.070 & -0.030\\
ZMR & 8.07 & 2.96 & 0.190 & 0.160 & 0.100 & 21.96 & 8.08 & 0.670 & 0.730 & \best{0.880}\\
B\&H & 51.65 & 16.93 & 0.690 & 0.750 & 0.460 & 36.36 & 12.90 & \second{0.700} & 0.930 & 0.480\\
\addlinespace[3pt]
\multicolumn{11}{@{}l}{\textit{RL-based Baselines}}\\
PPO & \second{54.70} & \second{17.35} & 0.658 & \second{0.802} & \second{0.527} & 19.28 & 6.99 & 0.327 & 0.486 & 0.221\\
A2C & 52.19 & 16.99 & \second{0.695} & 0.789 & 0.485 & 32.78 & 10.47 & 0.295 & 0.428 & 0.289\\
TD3 & 50.44 & 16.49 & 0.674 & 0.751 & 0.467 & 31.68 & 10.76 & 0.385 & 0.562 & 0.369\\
SAC & 36.78 & 12.33 & 0.480 & 0.552 & 0.320 & 19.47 & 6.41 & 0.242 & 0.385 & 0.181\\
DeepAries-k3 & 44.05 & 13.97 & 0.437 & 0.530 & 0.402 & 26.71 & 8.59 & 0.347 & 0.522 & 0.259\\
EarnMore & 51.53 & 16.65 & 0.570 & 0.709 & 0.435 & 13.80 & 4.76 & 0.186 & 0.306 & 0.149\\
DeepTrader & 8.90 & 2.27 & 0.120 & 0.269 & 0.194 & -6.98 & -3.35 & -0.182 & -0.235 & -0.019\\
MASA & 31.60 & 10.86 & 0.463 & 0.515 & 0.267 & \second{39.25} & \second{13.83} & \best{0.717} & \best{1.039} & 0.466\\
\addlinespace[3pt]
\multicolumn{11}{@{}l}{\textit{Our Method}}\\
Full FRPO & \best{75.32} & \best{22.78} & \best{0.796} & \best{0.986} & \best{0.620} & \best{48.50} & \best{15.67} & 0.624 & \second{0.991} & 0.550\\
\bottomrule
\end{tabular*}
}
\end{table*}

\subsection{Experimental Setup}
The experiments evaluate portfolio actions across markets and styles, then
examine relative decision supervision, predictive supervision, and their
coordination through component controls. Temporal splits and the fee sweep
test offline generalization and sensitivity to trading frictions.

\paragraph{Datasets.}
For the main benchmark, we train all methods from January 1, 2013, to December 31, 2018, and evaluate
them out of sample from January 1, 2019, to August 31, 2021. The evaluation
examines generalization along two complementary axes. First, DJIA30 and TOPIX
Core30 represent the U.S. and Japanese equity markets, respectively, and test
robustness across countries, market structures, and action-space scales.
Second, we construct three non-overlapping six-asset portfolios to evaluate
cross-style robustness under a fixed action dimension. The Mainstream
Technology and Indices portfolio contains QQQ, SPY, AAPL, AMZN, Alphabet, and
MSFT, following the canonical FinRL setting~\citep{liu2021finrl}. The Defensive
Blue-Chip portfolio contains BAC, JNJ, JPM, PG, WMT, and XOM, while the
High-Volatility Growth portfolio contains AVGO, BRK.B, META, NFLX, NVDA, and
TSLA. These portfolios differ in sector exposure, return distributions, and
risk profiles.

\paragraph{Implementation and Evaluation Protocol.}
FRPO uses 16-step observations, a two-layer Transformer encoder, group size
5, and Financial State weight $\alpha_0=4$. FRPO uses ten fixed
training seeds and 51,200 collected training transitions per seed. Training samples
stochastic actions; evaluation uses the deterministic mean.
Appendix~\ref{app:implementation} gives the full settings and checkpoint
schedule.

All methods, including the portfolio-specific baselines, are integrated into
the shared FinRL evaluation framework and tested with a uniform 1-bp
per-side transaction fee. Within each dataset, learned methods share the
asset panel, temporal split, trading calendar, initial capital, reward
scaling, daily decision frequency, and deterministic evaluation.
All methods receive identical asset
ordering, technical indicators, and portfolio states. The technical indicators are computed from
publicly available OHLCV data and require no proprietary financial signals.

\paragraph{Baselines and Metrics.}
The RL baselines are PPO~\citep{schulman2017ppo},
A2C~\citep{mnih2016asynchronous}, TD3~\citep{fujimoto2018td3}, and
SAC~\citep{haarnoja2018sac}. The portfolio-specific baselines are
DeepAries-k3~\citep{kim2025deeparies},
EarnMore~\citep{zhang2024earnmore},
DeepTrader~\citep{wang2021deeptrader}, and
MASA~\citep{li2024masa}. All baselines are reproduced from their official
GitHub repositories and undergo hyperparameter tuning in our experimental
setting. We report results using the tuned configurations.

We additionally evaluate deterministic MACD, SMA, ZMR, and buy-and-hold
(B\&H) strategies from January 2, 2019, to August 31, 2021, using price returns
without dividends and a 1-bp per-side cost. Each is a single backtest, so the
tables report point estimates. DIA and
1311.T serve as market proxies for DJIA30 and TOPIX Core30; Mainstream uses
SPY, while the other two B\&H portfolios hold 100 shares of each constituent.

We report cumulative return, annualized return, Sharpe ratio, Sortino ratio,
and Calmar ratio (Appendix~\ref{app:metrics}); higher values indicate better performance.

\subsection{Performance Evaluation across Markets and Investment Styles}

\begin{table*}[!t]
\caption{Performance on three heterogeneous six-asset portfolios. Reporting conventions follow Table~\ref{tab:large_markets}.}
\label{tab:heterogeneous_portfolios}
\centering
\resizebox{\linewidth}{!}{%
{
\fontsize{8}{9.6}\selectfont
\setlength{\tabcolsep}{3pt}
\renewcommand{\arraystretch}{1.08}
\begin{tabular}{l*{5}{r}@{\hspace{5pt}}*{5}{r}@{\hspace{5pt}}*{5}{r}}
\toprule
Method & \multicolumn{5}{c}{Mainstream Tech./Indices} & \multicolumn{5}{c}{Defensive Blue-Chip} & \multicolumn{5}{c}{High-Volatility Growth}\\
\cmidrule(lr){2-6} \cmidrule(lr){7-11} \cmidrule(lr){12-16}
 & CR (\%) & AR (\%) & Sharpe & Sortino & Calmar & CR (\%) & AR (\%) & Sharpe & Sortino & Calmar & CR (\%) & AR (\%) & Sharpe & Sortino & Calmar\\
\midrule
\multicolumn{16}{@{}l}{\textit{Technical and Market Baselines}}\\
MACD & 15.44 & 5.54 & 0.400 & 0.390 & 0.280 & 3.88 & 1.44 & 0.240 & 0.270 & 0.100 & 20.35 & 7.20 & 0.750 & 0.860 & 0.640\\
SMA & 56.27 & 18.25 & 0.950 & 1.010 & 0.790 & 6.05 & 2.23 & 0.340 & 0.370 & 0.170 & 39.16 & 13.21 & 1.020 & 1.130 & 0.850\\
ZMR & 39.53 & 13.33 & 0.770 & 0.800 & 0.490 & 10.88 & 3.96 & 0.240 & 0.280 & 0.130 & 66.49 & 21.10 & 1.090 & 1.200 & 0.730\\
B\&H & 80.48 & 24.82 & 1.090 & 1.180 & 0.730 & 17.70 & 6.31 & 0.630 & 0.750 & 0.390 & 62.47 & 19.99 & 1.440 & 1.720 & 1.240\\
\addlinespace[3pt]
\multicolumn{16}{@{}l}{\textit{RL-based Baselines}}\\
PPO & \second{159.54} & \second{42.87} & 1.598 & \second{1.978} & \second{1.585} & 55.74 & 18.02 & 0.728 & 0.920 & 0.596 & 208.08 & 49.47 & 1.423 & 1.838 & 1.405\\
A2C & 154.38 & 41.54 & 1.563 & 1.887 & 1.461 & 54.91 & 17.75 & 0.721 & 0.915 & 0.572 & 198.26 & 48.87 & 1.427 & 1.830 & 1.385\\
TD3 & 149.43 & 40.73 & 1.565 & 1.852 & 1.419 & 50.55 & 16.48 & 0.676 & 0.836 & 0.518 & 165.86 & 42.44 & 1.301 & 1.658 & 1.221\\
SAC & 125.19 & 35.54 & 1.467 & 1.649 & 1.156 & 32.73 & 11.08 & 0.491 & 0.610 & 0.359 & 144.43 & 36.73 & 1.104 & 1.414 & 1.089\\
DeepAries-k3 & 149.20 & 40.68 & \second{1.663} & 1.823 & 1.574 & 46.46 & 15.31 & 0.620 & 0.702 & 0.482 & \second{232.80} & \second{55.72} & \best{1.721} & \best{2.148} & \best{1.779}\\
EarnMore & 83.52 & 22.10 & 1.042 & 1.235 & 1.001 & 20.66 & 6.77 & 0.282 & 0.419 & 0.312 & 104.96 & 25.26 & 0.700 & 0.889 & 0.830\\
DeepTrader & -3.11 & -1.50 & -0.114 & -0.023 & 0.032 & 7.24 & 1.19 & 0.026 & 0.414 & 0.284 & -24.27 & -14.18 & -0.306 & -0.332 & -0.147\\
MASA & 152.29 & 41.56 & \best{1.673} & 1.955 & 1.541 & \second{58.18} & \second{18.79} & \second{0.837} & \second{1.035} & \second{0.608} & 160.04 & 43.18 & 1.494 & 1.693 & 1.147\\
\addlinespace[3pt]
\multicolumn{16}{@{}l}{\textit{Our Method}}\\
Full FRPO & \best{173.63} & \best{45.54} & 1.607 & \best{2.009} & \best{1.642} & \best{70.23} & \best{21.96} & \best{0.872} & \best{1.228} & \best{0.937} & \best{305.29} & \best{64.84} & \second{1.604} & \second{2.062} & \second{1.661}\\
\bottomrule
\end{tabular}
}}

\par\smallskip
\noindent
\begin{minipage}[t]{0.48\linewidth}
\vspace{0pt}
\centering
\captionsetup{type=figure,position=bottom}
\includegraphics[width=\linewidth]{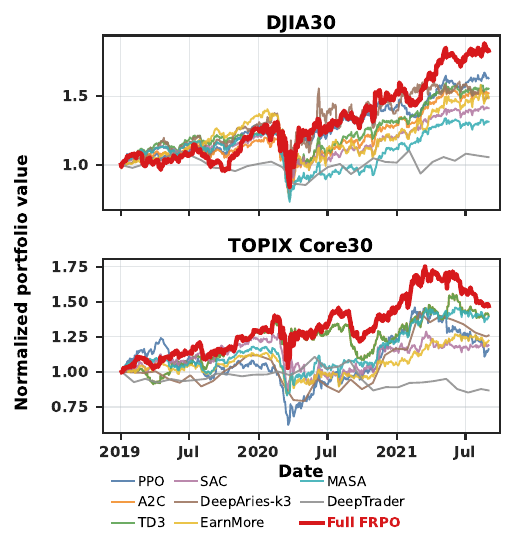}
\captionof{figure}{Representative portfolio values on DJIA30 and TOPIX Core30. Each curve uses the seed nearest its method's ten-seed mean; Full FRPO is thick red.}
\label{fig:representative_large_curves}
\Description{Representative portfolio-value trajectories on DJIA30 and TOPIX Core30.}
\end{minipage}\hfill
\begin{minipage}[t]{0.48\linewidth}
\vspace{0pt}
\centering
\captionsetup{type=figure,position=bottom}
\includegraphics[width=0.92\linewidth]{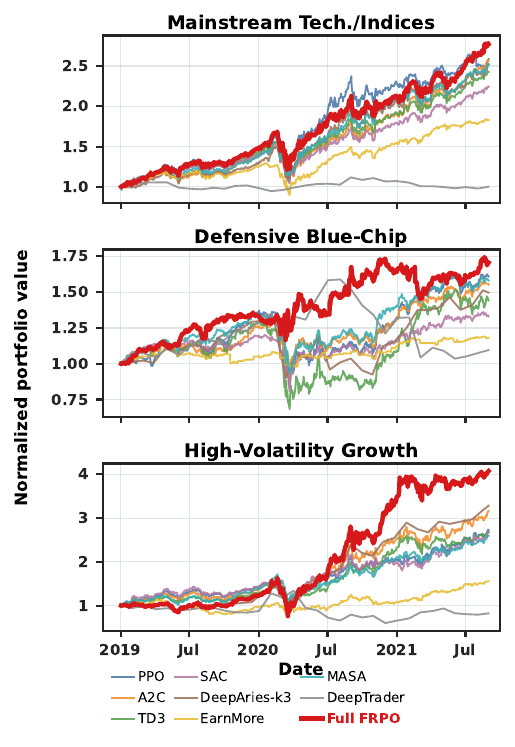}
\captionof{figure}{Representative portfolio values for the three six-asset portfolios. Seeds follow Figure~\ref{fig:representative_large_curves}; Full FRPO is thick red.}
\label{fig:representative_small_curves}
\Description{Representative portfolio-value trajectories on the Mainstream, Defensive, and High-Volatility Growth portfolios.}
\end{minipage}
\end{table*}

Tables~\ref{tab:large_markets} and~\ref{tab:heterogeneous_portfolios}
summarize performance across markets and investment styles, while
Figures~\ref{fig:representative_large_curves}
and~\ref{fig:representative_small_curves} show the corresponding representative
portfolio-value trajectories. Full FRPO achieves the highest cumulative and
annualized returns in all five datasets. Its cumulative returns are 75.32\%
and 48.50\% on DJIA30 and TOPIX Core30, respectively, and 173.63\%, 70.23\%,
and 305.29\% on the Mainstream, Defensive, and High-Volatility Growth portfolios.

Full FRPO also maintains competitive risk-adjusted performance. It attains the
best Sharpe, Sortino, and Calmar ratios on DJIA30 and the Defensive portfolio.
MASA leads Sharpe and Sortino on TOPIX Core30 and Sharpe on the Mainstream
portfolio, ZMR provides the highest TOPIX Calmar ratio, and DeepAries-k3 leads
the three risk-adjusted metrics on High-Volatility Growth.
Together, these results establish a consistent return advantage across the
five panels and competitive risk-adjusted performance. Component ablations
and same-path controls examine the auxiliary objectives and relative
learning design.

Across the 25 dataset--metric comparisons among learned methods (five datasets
$\times$ five metrics, excluding ablations), Full FRPO ranks first 19 times
and has the best average rank of 1.28 on these panels.

\subsection{Component Ablations}

\begin{table}[!htbp]
\centering
\footnotesize
\setlength{\tabcolsep}{3pt}
\caption{Component controls: ten-seed mean CR (\%). A: structural ablations on five panels; Gap is Full FRPO minus the stronger ablated variant. B: auxiliary weighting on two markets; Exponential and Fixed repeat Full and w/o Gate from A. Colors rank means within each market and panel.}
\label{tab:structural-ablation}
\label{tab:weighting_extension}
\begin{fitcolumn}
\begin{tabular}{lrrrr}
\toprule
\multicolumn{5}{l}{\textbf{A. Structural components}} \\
\midrule
Market & Full & w/o Gate & \shortstack{w/o Gate\\\& FSH} & Gap (pp) \\
\midrule
DJIA30 & \best{75.32} & \second{65.23} & 56.06 & +10.09 \\
TOPIX Core30 & \best{48.50} & \second{40.48} & 21.68 & +8.02 \\
Mainstream & \best{173.63} & \second{164.23} & 158.87 & +9.40 \\
Defensive & \best{70.23} & 56.08 & \second{57.39} & +12.84 \\
High-Vol. Growth & \best{305.29} & 214.61 & \second{224.51} & +80.78 \\
\bottomrule
\end{tabular}
\end{fitcolumn}
\par\smallskip
\begin{fitcolumn}
\begin{tabular}{lrrrr}
\toprule
\multicolumn{5}{l}{\textbf{B. Auxiliary weighting}} \\
\midrule
Market & Exponential & Fixed & Inverse & Kendall-style \\
\midrule
DJIA30 & $\best{75.32}$ & $\second{65.23}$ & $43.98$ & $57.18$ \\
TOPIX Core30 & $\best{48.50}$ & $\second{40.48}$ & $31.44$ & $19.54$ \\
\bottomrule
\end{tabular}
\end{fitcolumn}
\end{table}

\paragraph{Structural components.}
``FRPO w/o Gate'' retains the Financial State Head but replaces the adaptive
coefficient in Equation~\ref{eq:alphagate} with the fixed auxiliary weight
$\alpha_0=4$, disabling both variance and alignment gates;
``FRPO w/o Gate \& FSH'' removes both adaptive weighting and Financial State
supervision. Panel A of Table~\ref{tab:structural-ablation} compares future-state supervision and
its adaptive coordination with the policy objective. Full FRPO has higher mean cumulative return than either
variant in each of the five panels. Its margin over the stronger ablated
variant ranges from 8.02 to 80.78 percentage points. The comparison supports
the complete combination of relative decision supervision, forward-looking
representation learning, and adaptive coordination under the reported setups.

\paragraph{Auxiliary weighting.}
Table~\ref{tab:structural-ablation}B compares Exponential with Fixed,
Inverse, and Kendall-style weighting using ten paired seeds per large market.
Inverse changes the variance term while retaining alignment;
Kendall-style replaces both gates with a learned scale
(Appendix~\ref{app:aux-weighting}). Exponential has the highest mean CR
on both markets.

\subsection{Decision-Supervision Analysis}
\label{sec:decision-supervision}

\begin{table}[!htbp]
\centering
\footnotesize
\setlength{\tabcolsep}{3pt}
\caption{Same-path grouping on DJIA30 with FSH and Gate disabled: means over ten paired seeds. Higher CR, AR, and Sharpe and lower MDD are better. Colors follow Table~\ref{tab:large_markets}.}
\label{tab:samepath-ablation}
\begin{fitcolumn}
\begin{tabular}{lrrrr}
\toprule
Grouping & CR (\%) & AR (\%) & Sharpe & MDD (\%) \\
\midrule
Same-path & $\best{56.06}$ & $\best{18.13}$ & $\best{0.656}$ & $\best{39.03}$ \\
Independent windows & $\second{46.45}$ & $\second{15.05}$ & $\second{0.557}$ & $\second{40.41}$ \\
\bottomrule
\end{tabular}
\end{fitcolumn}
\end{table}

\paragraph{Same-path grouping.}
To examine grouping separately from auxiliary supervision, we disable
FSH and Adaptive Gate in both arms and match architecture, initialization,
and training budget across ten paired DJIA30 seeds. Independently sampled
candidate windows can overlap.
Table~\ref{tab:samepath-ablation} shows that same-path grouping increases
mean CR from $46.45\%$ to $56.06\%$, with higher AR and Sharpe and lower
MDD. The $9.61$-percentage-point mean paired return difference supports
comparing candidate decisions under a shared realized market window in
this configuration. Appendix~\ref{app:same-path-factor} analyzes the
dynamic factor-innovation risk.

\paragraph{Temporal credit assignment.}
We compare trajectory broadcasting with group-relative reward-to-go
(RTG) and immediate-reward signals, and with GAE, keeping the actor, FSH, Gate,
same-path sampling, and 51,200-transition budget fixed over ten seeds
per market. Returns are undiscounted. GAE uses a value head on detached
encoder features ($\gamma=1$, $\lambda=0.95$) with matched actor updates;
Appendix~\ref{app:credit-protocol} specifies the configurations.
Table~\ref{tab:credit-assignment-extended} shows that broadcasting has
the highest mean CR, AR, and Sharpe in both markets. Its mean CR exceeds
the strongest alternative by 11.58 percentage points on DJIA30 and
22.90 on TOPIX Core30. 

\begin{table}[!htbp]
\centering
\footnotesize
\setlength{\tabcolsep}{3pt}
\caption{Temporal credit assignment: ten-seed final-checkpoint means. Trajectory repeats Full FRPO; units and colors follow Table~\ref{tab:large_markets}.}
\label{tab:credit-assignment-extended}
\begin{fitcolumn}
\begin{tabular}{llrrr}
\toprule
Market & Estimator & CR (\%) & AR (\%) & Sharpe \\
\midrule
DJIA30 & Trajectory & \best{75.32} & \best{22.78} & \best{0.796} \\
 & RTG & \second{63.74} & \second{19.71} & \second{0.784} \\
 & Immediate & 46.89 & 15.49 & 0.631 \\
 & GAE & 41.43 & 13.79 & 0.582 \\
\midrule
TOPIX Core30 & Trajectory & \best{48.50} & \best{15.67} & \best{0.624} \\
 & RTG & 15.32 & 5.31 & 0.231 \\
 & Immediate & 18.02 & 6.26 & 0.227 \\
 & GAE & \second{25.60} & \second{9.12} & \second{0.399} \\
\bottomrule
\end{tabular}
\end{fitcolumn}
\end{table}

A supplementary PPO-GAE control retains the same auxiliary components
(Appendix~\ref{app:matched-ppo-fsh-gate}). The matched comparison supports
FRPO's complete training design in the tested configuration.

\begin{table*}[!t]
\centering
\fontsize{9}{10.8}\selectfont
\setlength{\tabcolsep}{3pt}
\renewcommand{\arraystretch}{1.18}
\caption{Temporal robustness: equal-weight means over the overlapping splits \texttt{w00}, \texttt{w01}, and \texttt{w02}, with ten seeds per split. Aggregation follows Equation~\ref{eq:three-window-mean}. CR and AR are percentages; higher is better. Colors follow Table~\ref{tab:large_markets}.}
\label{tab:temporal-summary-main}
\begin{tabular*}{\linewidth}{@{\extracolsep{\fill}}lrrrrr@{\hspace{12pt}}rrrrr@{}}
\toprule
 & \multicolumn{5}{c}{TOPIX Core30} & \multicolumn{5}{c}{DJIA30} \\
\cmidrule(lr){2-6}\cmidrule(lr){7-11}
Method & CR (\%) & AR (\%) & Sharpe & Sortino & Calmar & CR (\%) & AR (\%) & Sharpe & Sortino & Calmar \\
\midrule
FRPO & \best{50.52} & \best{15.76} & \second{0.592} & \second{0.923} & \best{0.524} & \best{64.93} & \best{19.59} & \best{0.695} & \best{0.855} & \best{0.535} \\
TD3 & 30.77 & 10.48 & 0.397 & 0.583 & 0.395 & 48.16 & 15.40 & \second{0.639} & \second{0.711} & 0.448 \\
SAC & 27.64 & 8.88 & 0.333 & 0.508 & 0.284 & 37.28 & 12.19 & 0.474 & 0.541 & 0.319 \\
A2C & 26.93 & 8.98 & 0.313 & 0.465 & 0.285 & 43.26 & 13.88 & 0.557 & 0.639 & 0.391 \\
PPO & 19.51 & 6.76 & 0.324 & 0.482 & 0.227 & \second{51.87} & \second{15.78} & 0.565 & 0.691 & \second{0.453} \\
EarnMore & 15.92 & 5.34 & 0.218 & 0.359 & 0.182 & 42.08 & 13.44 & 0.487 & 0.616 & 0.403 \\
DeepAries-k3 & 22.11 & 6.81 & 0.271 & 0.421 & 0.217 & 36.43 & 11.51 & 0.372 & 0.443 & 0.333 \\
MASA & \second{35.60} & \second{12.32} & \best{0.636} & \best{0.928} & \second{0.415} & 25.71 & 8.78 & 0.376 & 0.422 & 0.216 \\
\bottomrule
\end{tabular*}
\end{table*}

\subsection{Temporal Robustness}
\label{sec:temporal-robustness}
We evaluate FRPO and seven RL baselines on three overlapping temporal
splits: the main benchmark \texttt{w00} and two splits with later training
cutoffs and test start dates: \texttt{w01} tests from March 2019 to
November 2021, and \texttt{w02} from June 2019 to February 2022. Each method
is evaluated over ten training seeds per market and split. We average
metrics within each split and then give the three split means equal weight.
Appendix~\ref{app:temporal-extended} specifies all dates and individual results.

Table~\ref{tab:temporal-summary-main} shows that FRPO achieves the highest
mean CR and AR in both markets. Its mean CR is $50.52\%$ on TOPIX Core30
and $64.93\%$ on DJIA30, exceeding the strongest return baselines,
MASA and PPO, by $14.92$ and $13.06$ percentage points, respectively.
FRPO also leads mean Calmar in both markets and mean Sharpe and Sortino
on DJIA30; MASA leads mean Sharpe and Sortino on TOPIX Core30.
Appendix~\ref{app:temporal-extended} reports all split-level results.

\subsection{Transaction-Cost Sensitivity}
\begin{figure}[!htbp]
\centering
\includegraphics[width=\linewidth]{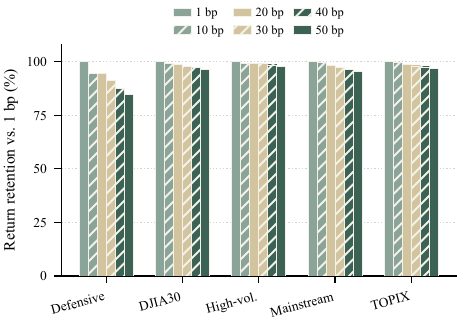}
\caption{Cumulative-return retention relative to the 1-bp setting as per-side
transaction costs increase across the five evaluation settings.}
\Description{Grouped bars show cumulative-return retention as per-side transaction costs increase from one to fifty basis points in five markets.}
\label{fig:transaction_cost_sensitivity}
\end{figure}

Figure~\ref{fig:transaction_cost_sensitivity} shows a smooth degradation as
per-side transaction costs increase from 1 to 50 bp. At 50 bp, FRPO retains
more than 95\% of its 1-bp return in four of the five settings; the Defensive
portfolio is more cost-sensitive but still retains 84.9\%. This result supports
robustness to substantially higher trading frictions while identifying the
Defensive setting as the one most affected by turnover costs.

Appendix~\ref{app:ablation-supplementary} provides metric definitions and weighting
details. Appendix~\ref{app:decision-diagnostics-extended} gives estimator
configurations and PPO diagnostics.

\FloatBarrier
\section{Conclusion}

Motivated by the needs for capital allocation and dynamic rebalancing in online investment services, this paper studies the opportunity-to-action problem in account-constrained portfolio decision learning. The proposed FRPO combines same-path relative supervision of trading trajectories with auxiliary learning of future cross-sectional opportunities, coordinated through a shared encoder and an Adaptive Gate. In experiments covering five portfolio settings, each evaluated with ten random seeds, FRPO achieves the highest mean cumulative and annualized returns among all compared methods. In addition, ablation studies, temporal split tests, and cost-sensitivity analyses further examine the performance of this design under different evaluation conditions.

\section*{Limitations}

The theoretical analysis in this paper focuses on the variance of centered factor innovations under non-anticipating trading policies, with the corresponding variance ordering holding in expectation over market and action randomness. Empirical evaluations are based on historical OHLCV data, technical indicators, and account states from 2019 to 2021, and include shifted temporal tests extending through February 2022. The current framework centers on structured financial signals and asset-level targets, with FSH variance characterizing uncertainty in financial targets. Future research could further incorporate multimodal Web information, end-to-end language agents, and user preference modeling.

\clearpage
\bibliographystyle{ACM-Reference-Format}
\begingroup
\def\showDOI#1{\mbox{#1}}
\bibliography{references}
\endgroup

\appendix
\section{Additional Evidence for FRPO}
\label{app:ablation-supplementary}
The experiments in this section use final checkpoints on the fixed
2019--2021 test history and the evaluation conventions of the main benchmark
unless a comparison specifies otherwise. Reported performance entries are
means across the specified training seeds.

\subsection{Implementation Details}
\label{app:implementation}
Full FRPO uses a 16-step state sequence and a shared two-layer Transformer
encoder (model dimension 128, four attention heads, feed-forward dimension
256), followed by LayerNorm(128), Linear(128,64), and GELU to return
64-dimensional features, with Financial State base weight $\alpha_0=4$. Main-benchmark settings
are group size 5, 1,024 rollout steps, seven update epochs, batch size 64,
learning rate $3\times10^{-4}$, clipping range 0.3, reward scaling $10^{-4}$,
and world-loss base 0. Training collects 51,200 transitions per seed over
ten policy iterations ($10\times5\times1{,}024$); checkpoints are saved every
25,000 steps. Each configuration uses ten fixed seeds. TOPIX retains four
update epochs and clipping range 0.2. Training uses stochastic policy samples;
evaluation uses deterministic mean actions. Experiments use an Intel Core
i9-13980HX CPU and an NVIDIA GeForce RTX 4090 GPU. The world-loss base is a
disabled implementation option, separate from the active FSH objective.

\subsection{Performance Metrics}
\label{app:metrics}
For evaluation, let $V_t$ be end-of-session NAV net of transaction costs,
$q_t=V_t/V_{t-1}-1$, and $T$ the number of daily returns; $K=252$.
For same-frequency risk-free returns $r_{f,t}$ and target return $m$,
write $e_t=q_t-r_{f,t}$ and $d_t=q_t-m$. Then
\begin{align*}
\mathrm{CR}&=V_T/V_0-1,&
\mathrm{AR}&=(V_T/V_0)^{K/T}-1,\\
\mathrm{Sharpe}&=\sqrt K\,\bar e/s_e,&
\mathrm{Sortino}&=\frac{\sqrt K\,\bar d}
 {\sqrt{T^{-1}\sum_{t=1}^T\min(d_t,0)^2}}.
\end{align*}
Here $s_e^2=(T-1)^{-1}\sum_t(e_t-\bar e)^2$; downside deviation
includes all $T$ sessions. With positive maximum drawdown,
\begin{align*}
\mathrm{MDD}&=\max_{0\le t\le T}
 \left(1-\frac{V_t}{\max_{0\le s\le t}V_s}\right),&
\mathrm{Calmar}&=\mathrm{AR}/\mathrm{MDD}.
\end{align*}
CR and AR are reported in percent; ratios use unscaled returns.

\subsection{Auxiliary Objective Weighting}
\label{app:aux-weighting}
We compare the original exponential weighting with three alternatives on
DJIA30 and TOPIX Core30 using ten fixed paired seeds per market
(10 seeds $\times$ 2 markets). Fixed uses
$\alpha_0=4$ without the uncertainty or alignment gates. Inverse replaces
$\exp(-u/\tau)$ with $\tau/(u+10^{-8})$ while retaining alignment;
Kendall-style uses a learned loss scale in place of both gates. Table~\ref{tab:weighting_extension}, panel B, reports the results in the main text.

\section{Temporal Robustness}
\label{app:temporal-extended}
\subsection{Evaluation across Shifted Temporal Splits}
\label{app:shifted-splits-extended}
We compare FRPO with seven baselines on TOPIX Core30 and DJIA30 across
the main split \texttt{w00} and the shifted splits \texttt{w01}/\texttt{w02}.

\paragraph{Training and test intervals.}
Table~\ref{tab:temporal-date-splits} gives inclusive calendar boundaries
shared by both markets; testing uses the available trading sessions.
All methods follow the FinRL protocol with a 1-bp per-side fee and ten
training seeds per market and split.

\begin{table}[t]
\centering
\footnotesize
\setlength{\tabcolsep}{3pt}
\renewcommand{\arraystretch}{1.15}
\caption{Inclusive training/test calendar boundaries shared by TOPIX Core30 and DJIA30.}
\label{tab:temporal-date-splits}
\begin{fitcolumn}
\begin{tabular}{lll}
\toprule
Date split & Train (both markets) & Test (both markets) \\
\midrule
\texttt{w00} & 2013-01-01 $\sim$ 2018-12-31 & 2019-01-01 $\sim$ 2021-08-31 \\
\texttt{w01} & 2013-03-01 $\sim$ 2019-02-28 & 2019-03-01 $\sim$ 2021-11-30 \\
\texttt{w02} & 2013-06-01 $\sim$ 2019-05-31 & 2019-06-01 $\sim$ 2022-02-28 \\
\bottomrule
\end{tabular}
\end{fitcolumn}
\end{table}

\paragraph{Performance across individual splits.}
Table~\ref{tab:temporal-split-results} reports the shifted splits;
Table~\ref{tab:large_markets} gives \texttt{w00} results.

\begin{table*}[t]
\centering
\fontsize{8.5}{10.2}\selectfont
\setlength{\tabcolsep}{3pt}
\renewcommand{\arraystretch}{1.10}
\caption{Ten-seed mean performance on shifted splits \texttt{w01} and
\texttt{w02}; dates follow Table~\ref{tab:temporal-date-splits}, and
\texttt{w00} results appear in Table~\ref{tab:large_markets}.
CR and AR are percentages; higher is better. \best{Red}/\second{blue} mark
the best/second-best means per split, market, and metric.}
\label{tab:temporal-split-results}
\begin{fitcolumn}
\begin{tabular*}{\linewidth}{@{\extracolsep{\fill}}l*{5}{r}@{\hspace{10pt}}*{5}{r}@{}}
\toprule
Method & \multicolumn{5}{c}{TOPIX Core30} & \multicolumn{5}{c}{DJIA30} \\
\cmidrule(lr){2-6}\cmidrule(lr){7-11}
 & CR (\%) & AR (\%) & Sharpe & Sortino & Calmar & CR (\%) & AR (\%) & Sharpe & Sortino & Calmar \\
\midrule
\multicolumn{11}{l}{\textbf{w01}} \\
\addlinespace[2pt]
FRPO & \best{55.18} & \best{16.68} & \best{0.593} & \best{0.937} & \best{0.531} & \best{53.26} & \best{16.19} & \best{0.639} & \best{0.761} & \best{0.496} \\
TD3 & \second{28.89} & \second{9.99} & 0.372 & 0.553 & \second{0.370} & \second{42.85} & \second{13.50} & \second{0.570} & \second{0.627} & \second{0.408} \\
SAC & 22.86 & 7.53 & 0.275 & 0.398 & 0.235 & 30.77 & 9.99 & 0.393 & 0.447 & 0.267 \\
A2C & 21.22 & 7.39 & 0.272 & 0.409 & 0.230 & 35.81 & 11.32 & 0.440 & 0.504 & 0.310 \\
PPO & 12.37 & 4.08 & 0.199 & 0.306 & 0.147 & 40.32 & 11.51 & 0.418 & 0.488 & 0.336 \\
EarnMore & 13.59 & 4.69 & 0.206 & 0.322 & 0.161 & 28.87 & 9.20 & 0.374 & 0.455 & 0.316 \\
DeepAries-k3 & 21.27 & 6.26 & 0.244 & 0.393 & 0.207 & 35.04 & 10.90 & 0.343 & 0.411 & 0.307 \\
MASA & 25.94 & 9.10 & \second{0.471} & \second{0.690} & 0.307 & 18.12 & 6.23 & 0.269 & 0.303 & 0.153 \\
\midrule
\multicolumn{11}{l}{\textbf{w02}} \\
\addlinespace[2pt]
FRPO & \best{47.88} & \best{14.93} & \second{0.559} & \second{0.842} & \best{0.490} & \best{66.21} & \best{19.80} & \second{0.650} & \best{0.818} & \second{0.489} \\
TD3 & 31.75 & 10.70 & 0.433 & 0.635 & 0.447 & 51.19 & 16.20 & \best{0.672} & 0.756 & 0.469 \\
SAC & 40.58 & 12.69 & 0.482 & 0.741 & 0.437 & 44.28 & 14.24 & 0.550 & 0.625 & 0.369 \\
A2C & 26.78 & 9.08 & 0.373 & 0.559 & 0.336 & 41.79 & 13.33 & 0.537 & 0.623 & 0.377 \\
PPO & 26.89 & 9.21 & 0.445 & 0.654 & 0.314 & \second{60.60} & \second{18.47} & 0.620 & \second{0.783} & \best{0.496} \\
EarnMore & 20.36 & 6.57 & 0.263 & 0.449 & 0.236 & 45.85 & 14.47 & 0.517 & 0.684 & 0.458 \\
DeepAries-k3 & 18.35 & 5.59 & 0.221 & 0.347 & 0.184 & 30.19 & 9.66 & 0.335 & 0.389 & 0.289 \\
MASA & \second{41.60} & \second{14.04} & \best{0.719} & \best{1.055} & \second{0.473} & 27.42 & 9.24 & 0.396 & 0.447 & 0.227 \\
\bottomrule
\end{tabular*}
\end{fitcolumn}
\end{table*}

FRPO leads mean CR and AR in all six market--split comparisons.
Risk-adjusted leaders vary: in \texttt{w02}, TD3 leads Sharpe and PPO
leads Calmar on DJIA30, while MASA leads Sharpe and Sortino on TOPIX Core30.

\paragraph{Aggregation across temporal splits.}
For each method and market, let $m_{w,s}$ denote a metric measured on
split $w$ under training seed $s$. For $\mathcal W=\{\texttt{w00},\texttt{w01},\texttt{w02}\}$, we report the equal-weight mean
\begin{equation}
\bar m = \frac{1}{3}\sum_{w\in\mathcal W}
\left(\frac{1}{10}\sum_{s=1}^{10} m_{w,s}\right).
\label{eq:three-window-mean}
\end{equation}
Each split receives equal weight after averaging its ten seeds.
Sharpe, Sortino, and Calmar are means of split-level metrics, not metrics
recomputed from a concatenated return series.

Table~\ref{tab:temporal-summary-main} reports all five aggregate metrics in the main text.

\section{Supplementary Decision-Supervision Evidence}
\label{app:decision-diagnostics-extended}

\subsection{Temporal Credit Assignment}
\label{app:credit-protocol}
We compare trajectory broadcasting, group-relative reward-to-go (RTG),
group-relative immediate rewards, and GAE, holding the actor, FSH, Gate,
same-path sampling, and update budget fixed. All use FinRL, a 1-bp per-side
fee, and final 51,200-step checkpoints. DJIA30 uses seven epochs and
clipping $0.3$; TOPIX uses four epochs and clipping $0.2$.
GAE fits a 4,225-parameter MLP on detached 64-dimensional encoder features to
$\lambda$-returns ($\gamma=1$, $\lambda=0.95$), without encoder
backpropagation and with matched actor updates. All returns are
undiscounted. Table~\ref{tab:credit-assignment-extended} in the main text reports the results.

\subsection{PPO Value-Prediction and Training Diagnostics}
For twenty final PPO checkpoints (ten seeds per market), we save a
reproducible stochastic trajectory over 2019--2021 in FinRL with a 1-bp
per-side fee. Pre-action value predictions are compared with realized
reward-to-go using the original reward, $\gamma=0.99$, and zero terminal
continuation. Error includes prediction and sampled-return noise.
High/low regimes use the upper/lower quartiles of trailing 20-session
equal-weight market volatility, fitted on training data and fixed on test
dates. High-regime MSE is about $2.1$/$2.2$ times low-regime MSE on
DJIA30/TOPIX; explained variance and correlation show different patterns
(Table~\ref{tab:ppo-value-diagnostics-extended}, panel A).
The same runs provide six online value-loss/EV snapshots per seed
(120 total); panel B shows the first and last, at 8,192 and 49,152
logged transitions within the 51,200-transition run. Online targets
are rollout $\lambda$-returns; test targets are realized discounted
reward-to-go, so their errors are not directly comparable.

\begin{table}[t]
\centering
\footnotesize
\setlength{\tabcolsep}{3pt}
\caption{PPO diagnostics, ten-seed means per market. A: final-checkpoint test predictions; MSE uses squared reward units, EV is $1-\mathrm{Var}(R-V)/\mathrm{Var}(R)$, and $r$ is Pearson correlation with realized reward-to-go. B: online training snapshots; training ends at 51,200 transitions. Targets differ between panels.}
\label{tab:ppo-value-diagnostics-extended}
\label{tab:ppo-value-training-extended}
\begin{fitcolumn}
\begin{tabular}{llrrr}
\toprule
\multicolumn{5}{l}{\textbf{A. Test reward-to-go prediction}} \\
\midrule
Market & Test regime & Value MSE & EV & $r$ \\
\midrule
DJIA30 & All & 1.3086 & $-0.2204$ & 0.2995 \\
 & High & 1.6317 & $-0.4671$ & 0.2364 \\
 & Low & 0.7723 & $-2.2793$ & 0.4170 \\
\midrule
TOPIX Core30 & All & 1.7173 & $-0.0210$ & 0.0320 \\
 & High & 2.7314 & $-0.0170$ & 0.1219 \\
 & Low & 1.2236 & $-0.0122$ & $-0.0387$ \\
\midrule
\multicolumn{5}{l}{\textbf{B. Online rollout-target fitting}} \\
\midrule
Market & \shortstack{Logged\\transitions} & Value loss & EV & \\
\midrule
DJIA30 & 8,192 & 0.096 & 0.027 & \\
 & 49,152 & 0.164 & 0.466 & \\
\midrule
TOPIX Core30 & 8,192 & 0.456 & 0.116 & \\
 & 49,152 & 0.585 & 0.092 & \\
\bottomrule
\end{tabular}
\end{fitcolumn}
\end{table}

\subsection{Matched PPO-GAE Control with Shared Auxiliary Components}
\label{app:matched-ppo-fsh-gate}

To assess the complete FRPO training design while retaining the same
auxiliary prediction and weighting mechanisms, we compare Full FRPO with a
matched PPO-GAE variant that retains FSH and Gate.

The DJIA30 ten-seed control in Table~\ref{tab:matched-ppo-fsh-gate}
retains FRPO's actor, encoder, FSH, Gate, and optimization configuration,
replacing group-relative advantages with GAE. Final checkpoints use 51,200
collected training transitions per seed over ten iterations.
FinRL evaluation uses 1-bp per-side fees and
2019-01-02--2021-08-31.

The added 64--64--1 Tanh value MLP takes the 64-dimensional encoder features
and has 4,225 parameters. Its loss
backpropagates through the encoder; actor, FSH, and value head share Adam.
GAE uses $\gamma=0.99$, $\lambda=0.95$, zero terminal continuation,
and normalization across five concatenated rollouts. Value targets are
unnormalized GAE plus rollout value predictions; the objective adds
$0.5$ times value MSE. Auxiliary weighting follows
Equation~\ref{eq:alphagate}, with gradient alignment measured between the
policy loss and the unweighted FSH loss on shared encoder sequence features.
Settings are rollout length 1,024, five trajectories, batch size 64,
seven epochs, learning rate $3\times10^{-4}$, and clipping $0.3$.

\begin{table}[!htbp]
\centering
\footnotesize
\setlength{\tabcolsep}{3pt}
\renewcommand{\arraystretch}{1.08}
\caption{Supplementary DJIA30 control with matched actor and encoder architectures, FSH, and Gate. Ten-seed means at the final 51,200-step checkpoint; Full FRPO repeats Table~\ref{tab:large_markets}, and units and colors follow that table.}
\label{tab:matched-ppo-fsh-gate}
\begin{fitcolumn}
\begin{tabular}{lrrr}
\toprule
Training framework & CR (\%) & AR (\%) & Sharpe \\
\midrule
Full FRPO & \best{75.32} & \best{22.78} & \best{0.796} \\
PPO-GAE + FSH + Gate & \second{53.35} & \second{15.95} & \second{0.545} \\
\bottomrule
\end{tabular}
\end{fitcolumn}
\end{table}

Table~\ref{tab:matched-ppo-fsh-gate} reports higher mean CR, AR, and Sharpe
for Full FRPO. With the auxiliary components present in both arms, these
results provide supplementary evidence for the complete decision-supervision
and update scheme in the tested configuration. The two schemes differ
jointly in advantage construction, discounting, normalization, and shared
critic gradients.

\section{Dynamic Factor-Noise Analysis of Same-Path Comparison}
\label{app:same-path-factor}

We extend the $\mathcal C,\mathcal D$ decomposition to dynamic accounts,
analyzing centered factor innovations before group standardization.

\paragraph{Account timing and return units.}
Let $V_{g,t}>0$ be pre-trade wealth, $\mathbf h^+_{g,t}$ holdings
executed at prices $\mathbf p_t$, and $K_{g,t}$ fees. With interest-free
cash and no external flows, trade, deduct fees, then advance prices:

\begin{equation}
V_{g,t+1}-V_{g,t}
=(\mathbf p_t\odot\mathbf h^+_{g,t})^\top\mathbf r_{t+1}-K_{g,t}.
\label{eq:account-dynamic}
\end{equation}
Here $\mathbf r_{t+1}$ denotes simple asset returns. Training uses
$r_t^{(g)}=c\Delta V_{g,t}$, where $c=10^{-4}$ and
$\Delta V_{g,t}=V_{g,t+1}-V_{g,t}$ includes fees. Set
$\mathbf x_{g,t}=c(\mathbf p_t\odot\mathbf h^+_{g,t})$;
then $R_g=\sum_t r_t^{(g)}=c(V_{g,T}-V_0)$ with $\gamma=1$.
Holdings reflect cash limits, integer fills, fees, and feedback.
Equal portfolio proportions can give different factor P\&L at different wealth.

\paragraph{Predictable dynamic exposures.}
Fix policy parameters within a group. The joint filtration
$(\mathcal F_t)$ contains market history and all candidates' accounts,
executed trading actions, and action draws through $t$, but no future information.
For $\mathcal F_{t+1}$-measurable innovations, assume

\begin{equation}
\begin{aligned}
\mathbf r_{t+1}&=\boldsymbol\mu_t+B_t\boldsymbol\varepsilon_{t+1}
                  +\boldsymbol\xi_{t+1},\\
\mathbb E[\boldsymbol\varepsilon_{t+1}\mid\mathcal F_t]&=0,\qquad
\operatorname{Cov}(\boldsymbol\varepsilon_{t+1}\mid\mathcal F_t)=\Sigma_t.
\end{aligned}
\label{eq:factor-model}
\end{equation}
The exogenous $B_t,\Sigma_t$ and exposures $\mathbf x_{g,t}$ are
$\mathcal F_t$-measurable, allowing feedback and conditional
heteroscedasticity. Predictable raw-factor components enter
$\boldsymbol\mu_t$. Define

\begin{equation}
\mathbf b_{g,t}=B_t^\top\mathbf x_{g,t},\qquad
M_g=\sum_t\mathbf b_{g,t}^\top\boldsymbol\varepsilon_{t+1},
\quad R_g=U_g+M_g.
\label{eq:factor-trajectory}
\end{equation}
Require finite energy
$\mathbb E\sum_t\mathbf b_{g,t}^\top\Sigma_t\mathbf b_{g,t}<\infty$.
$U_g$ collects drift, fees, and other returns; it may covary with $M_g$.

\paragraph{Dynamic variance decomposition.}
Set $\bar{\mathbf b}_t=G^{-1}\sum_g\mathbf b_{g,t}$ and
$\boldsymbol\delta_{g,t}=\mathbf b_{g,t}-\bar{\mathbf b}_t$. Define
\begin{equation}
\begin{aligned}
\mathcal C_{\rm dyn}
 &=\mathbb E\sum_t\bar{\mathbf b}_t^\top\Sigma_t\bar{\mathbf b}_t,\\
\mathcal D_{\rm dyn}
 &=\mathbb E\sum_t\frac1G\sum_g
       \boldsymbol\delta_{g,t}^\top\Sigma_t\boldsymbol\delta_{g,t}.
\end{aligned}
\label{eq:factor-dispersion}
\end{equation}
Expectations cover market and action randomness with a fixed initial
account and starting market condition. Define
$V_{\rm same}=G^{-1}\sum_g\operatorname{Var}(M_g-\bar M)$,
$\bar M=G^{-1}\sum_gM_g$. Independently retraded paths rebuild accounts
and exposures from independent market/action draws, matching each
candidate's same-path trajectory marginal. Define $V_{\rm ind}$ analogously.
For $G\geq2$,

\begin{equation}
\begin{aligned}
V_{\rm same}&=\mathcal D_{\rm dyn},\\
V_{\rm ind}&=\left(1-\frac1G\right)
                 (\mathcal C_{\rm dyn}+\mathcal D_{\rm dyn}),\\
V_{\rm ind}-V_{\rm same}
 &=\frac{(G-1)\mathcal C_{\rm dyn}-\mathcal D_{\rm dyn}}{G}.
\end{aligned}
\label{eq:factor-variance-comparison}
\end{equation}
Lower average innovation variance holds exactly when
$\mathcal D_{\rm dyn}<(G-1)\mathcal C_{\rm dyn}$, in expectation
rather than for every realized group.

\paragraph{Proof.}
Conditional zero mean gives $\mathbb E[M_g]=0$; predictable exposures
make different-time martingale increments orthogonal. Thus

\begin{equation}
\begin{aligned}
M_g-\bar M&=\sum_t
 \boldsymbol\delta_{g,t}^\top\boldsymbol\varepsilon_{t+1},\\
\operatorname{Var}(M_g-\bar M)
 &=\mathbb E\sum_t
 \boldsymbol\delta_{g,t}^\top\Sigma_t\boldsymbol\delta_{g,t}.
\end{aligned}
\label{eq:factor-centered}
\end{equation}
Averaging gives $V_{\rm same}=\mathcal D_{\rm dyn}$.
Let $q_g=\mathbb E\sum_t\mathbf b_{g,t}^\top\Sigma_t\mathbf b_{g,t}
=\operatorname{Var}(M_g)$, also the variance of its matched independent
counterpart $\widetilde M_g$. Independence and zero means give
\begin{equation}
\begin{aligned}
\operatorname{Var}(\widetilde M_g-\overline{\widetilde M})
 &=\left(1-\frac1G\right)^2q_g+\frac1{G^2}\sum_{h\ne g}q_h,\\
\frac1G\sum_gq_g&=\mathcal C_{\rm dyn}+\mathcal D_{\rm dyn}.
\end{aligned}
\label{eq:factor-independent}
\end{equation}
The second identity uses $\mathbf b_{g,t}=\bar{\mathbf b}_t+
\boldsymbol\delta_{g,t}$ and $\sum_g\boldsymbol\delta_{g,t}=0$.
Averaging the first proves Equation~\ref{eq:factor-variance-comparison}.

\paragraph{Residual factor risk between candidates.}
For $Q_t=B_t\Sigma_tB_t^\top$, the pairwise dispersion identity gives
\begin{equation}
\mathcal D_{\rm dyn}=\mathbb E\sum_t\frac1{2G^2}\sum_{g,h}
(\mathbf x_{g,t}-\mathbf x_{h,t})^\top Q_t
(\mathbf x_{g,t}-\mathbf x_{h,t}).
\label{eq:factor-pairwise-risk}
\end{equation}
This is dynamic active factor risk. Equal-beta stocks can cancel a market
innovation despite different trades if effective exposures match.
Differences in beta, cash allocation, or account wealth can leave residuals;
a shared pool, long-only trading, and no leverage do not ensure cancellation.
Zero $\Sigma_t$-seminorm dispersion cancels the innovation; generally
$V_{\rm same}\leq\mathbb E\sum_t\lambda_{\max}(\Sigma_t)
G^{-1}\sum_g\|\boldsymbol\delta_{g,t}\|_2^2$.
Any identical additive $M$ cancels exactly:
$R_g=U_g+M$ implies $R_g-\bar R=U_g-\bar U$.

\paragraph{Relation to fixed exposures.}
For fixed stacked exposures $\boldsymbol\beta_g$, factor P\&L is
$\boldsymbol\beta_g^\top\mathbf F$ with $\mathbb E[\mathbf F]=0$,
$\operatorname{Cov}(\mathbf F)=\Sigma$. Set
$\mathcal C=\bar{\boldsymbol\beta}^\top\Sigma\bar{\boldsymbol\beta}$,
$\mathcal D=G^{-1}\sum_g
(\boldsymbol\beta_g-\bar{\boldsymbol\beta})^\top\Sigma
(\boldsymbol\beta_g-\bar{\boldsymbol\beta})$.
The three identities still hold, allowing arbitrary cross-time covariance.
The dynamic extension instead requires conditional-zero-mean innovations.
Fees and other returns may covary with $M_g$; neither model guarantees
lower total-return or standardized policy-gradient variance.

\section{Open-Source Resources}
\label{app:open-source-resources}
We will release (1) FRPO code for the core algorithm, training, and evaluation;
(2) configurations for data processing, models, and training hyperparameters;
(3) scripts reproducing main experiments, baseline comparisons, and ablations;
and (4) analysis and visualization scripts for aggregating results and
generating the paper's tables and figures.
The README will document data sources, environment dependencies, and
execution commands to support reproduction and further research.

\end{document}